\documentclass[lettersize,journal]{IEEEtran}
\usepackage{amsmath,amsfonts}
\usepackage{algorithmic}
\usepackage{algorithm}
\usepackage{array}
\usepackage[caption=false,font=normalsize,labelfont=sf,textfont=sf]{subfig}
\usepackage{textcomp}
\usepackage{stfloats}
\usepackage{url}
\usepackage{verbatim}
\usepackage{graphicx}
\usepackage{cite}
\usepackage{booktabs}
\usepackage{tabularx}
\usepackage{multirow}
\usepackage{hyperref}
\usepackage[shortlabels]{enumitem}
\begin{document}
	\title{A Data-Driven Modeling and Motion Control of Heavy-Load Hydraulic Manipulators via Reversible Transformation}
	\author{Dexian Ma, Yirong Liu, Wenbo Liu, and Bo Zhou, \IEEEmembership{Member, IEEE}
		\thanks{This work was supported in part by the National Natural Science Foundation (NNSF) of China under Grant 62073075. \textit{(Corresponding author: Bo Zhou.)}}
		\thanks{D. X. Ma, Y. R. Liu, W. B. Liu, and B. Zhou are with the Key Laboratory of Measurement and Control of Complex Systems of Engineering (School of Automation, Southeast University), Ministry of Education, Nanjing 210096, China (e-mail: {mdx@seu.edu.cn}, {liu\_yirong@seu.edu.cn}, {lwb@seu.edu.cn}, {zhoubo@seu.edu.cn},). }
	}

	\maketitle
	
	\begin{abstract}
		This work proposes a data-driven modeling and the corresponding hybrid motion control framework for unmanned and automated operation of industrial heavy-load hydraulic manipulator. Rather than the direct use of a neural network black box, we construct a reversible nonlinear model by using multilayer perceptron to approximate dynamics in the physical integrator chain system after reversible transformations. The reversible nonlinear model is trained offline using supervised learning techniques, and the data are obtained from simulations or experiments. Entire hybrid motion control framework consists of the model inversion controller that compensates for the nonlinear dynamics and proportional-derivative controller that enhances the robustness. The stability is proved with Lyapunov theory. Co-simulation and Experiments show the effectiveness of proposed modeling and hybrid control framework. With a commercial 39-ton class hydraulic excavator for motion control tasks, the root mean square error of trajectory tracking error decreases by at least 50\% compared to traditional control methods. In addition, by analyzing the system model, the proposed framework can be rapidly applied to different control plants.
	\end{abstract}
	
	
	\section{Introduction}\label{introduction}
	\IEEEPARstart{I}{n} some major engineering tasks such as construction, mining, and waste disposal, where there is a demand for heavy load (at least reaching ton level), high precision, and high speed, hydraulic manipulators have become the preferred choice due to the exceptional power-to-weight ratio. Considering economic costs and application requirements, an interesting approach is to carry out the unmanned transformation of existing hydraulic excavators, evolving them into automated mobile manipulator systems \cite{b1}. However, most commercial hydraulic excavators are designed for manual control, with hydraulic circuits and mechanical systems that are not finely engineered, making the unmanned retrofitting and automatic control a challenging issue.
	
	Currently, the methodologies for controlling hydraulic manipulators are primarily categorized into model-based approaches and data-driven approaches. Model-based approaches rely upon the utilization of mechanistic robotic models. Some methods have been proposed for hydraulic manipulators \cite{b2},\cite{b3},\cite{b4}. However, it is typically difficult to acquire open-source information from manufacturers, and the imprecise design further complicates the task of describing it with an accurate mathematical model. Another challenge comes from the complex nonlinearity of the hydraulic manipulator system such as fluid compressibility, the dead-zone and hysteresis effects of servo valves, actuator leakage, and the interaction between several actuators that share a common supply pump \cite{b5}. Obtaining all the mathematical model parameters is difficult in engineering, and the simplified mathematical model faces challenges in accurately control.
	
	Data-driven approaches utilize offline or online data from plants for modeling and control. A classical method is to use neural networks, which have the capability to approximate any nonlinear function \cite{b6},\cite{b7}. Neural networks offer a valuable method for modeling nonlinear dynamic systems that comprise: 1) a forward model that predicts the future states based on input signal and current states, 2) an inverse model that derives the input signal from the current state and target state. Researches indicate that data-driven controller (DDC) based on neural networks have been employed in hydraulic manipulator systems, and the main approach is to utilize the neural network to approximate forward and inverse dynamics. The forward model can be utilized for model predictive control \cite{b8},\cite{b9}, reinforcement learning \cite{b10}, model referencing adaptive control \cite{b11}, state observer \cite{b12} and the inverse model can be used for direct control \cite{b13},\cite{b14} or feedforward compensation \cite{b15}. Data-driven method can also combine with prior knowledge to improve the control performance. Lee et al. \cite{b16} developed a physics-inspired data-driven model with a modular architecture, comprising dead-time, time delay, dynamics and inverse dynamics, which has improved operational accuracy. Jonas et al. \cite{b17} combined data-driven control with expert knowledge, enhancing the ability to handle external loads. 
	
	The application of data-driven approaches remains largely unexplored within current landscape. The forward and inversion dynamics are often constructed as pure black-box models in DDC, neglecting the inherent physical characteristics, which leads to difficulties in designing controllers and poor interpretability. Furthermore, due to the physical infeasibility of directly obtaining inverse models through the inversion of forward dynamics, the forward and inverse models required for DDC must be established separately, which cannot ensure their correlation \cite{b18}. Another issue is that pure black-box model lacks theoretical proofs of stability, which leads to uncertainty of control performance and makes it difficult to generalize.
	
	According to the above ideas and deficiencies, we aim to design universal modeling and control methods for hydraulic manipulators to achieve automation operation. The contributions of this work are as follows:
	
	\begin{enumerate}
		\item A physics guided reversible nonlinear model (RevNM) is proposed. Different from the pure black-box modeling approach, an analysis of the hydraulic manipulator's nonlinear dynamical model which includes dead zones, hysteresis, and leakage has been conducted, culminating in the formulation of its third-order differential equation. Based on this prior information, the third-order data-driven model has been constructed.
		\item By employing reversible transformations, both the forward and inverse models of the RevNM can be obtained simultaneously. It should be noticed that the RevNM can be extended to general nonlinear integrator chain systems with prior analysis.
		\item A corresponding hybrid control framework that incorporates model inversion controller and proportional-derivative (PD) controller has been constructed. The model inversion controller compensates for nonlinearities, and the PD controller enhances the robustness. 
		\item The proposed RevNM and hybrid control framework are demonstrated to be effective by co-simulation and experimental tests on a 39-ton commercial hydraulic excavator. Within our framework, the tracking performance improves at least 50\% compared to traditional methods.
	\end{enumerate}
	
	\section{System Description}\label{system}
\begin{figure}[htbp]
		\centerline{\includegraphics[width=\columnwidth]{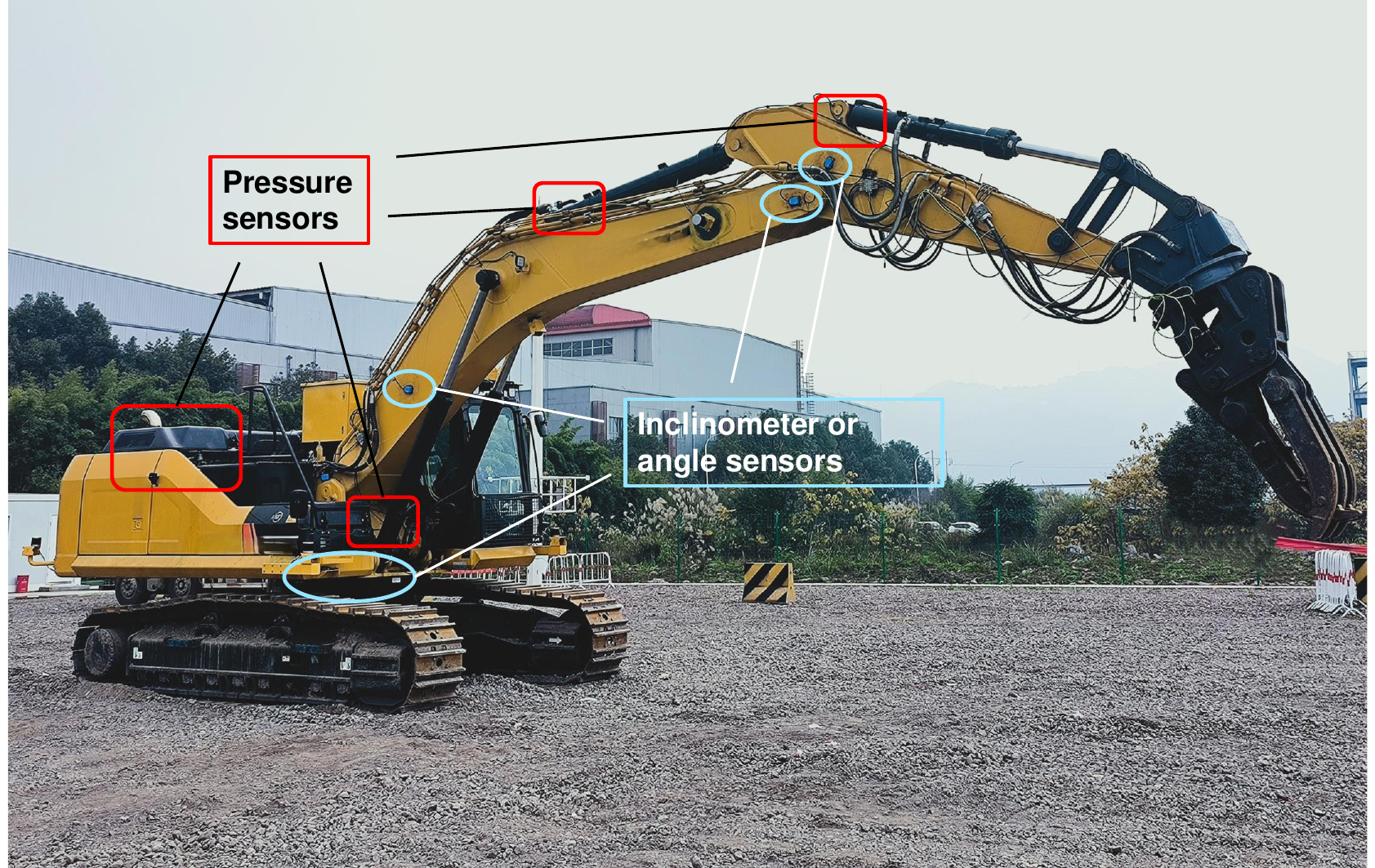}}
		\caption{A commercial 39-ton class industrial hydraulic excavator. The excavator is customized with inclinometer sensors, angle sensors, and pressure sensors.}
		\label{excavator}
	\end{figure}

	The commercially 39-ton hydraulic excavator consists of engines, variable-displacement piston pumps, main control valves, hydraulic actuators, hydraulic motors, mechanical arms, and the cab. The boom, arm, and bucket linkage are equipped with inclination sensors, while the swing angle can be measured through built-in angular sensors. Both the hydraulic actuators and hydraulic motors are fitted with pressure sensors to monitor the state of the excavator, as shown in Fig.\ref{excavator}. We implement remote control based on the user datagram protocol (UDP) communication, using two computers termed as the transmission and control units. The transmission computer is located in the cab of the excavator and is responsible for transmitting state information back to the controller computer. It also relays the valve spool control commands calculated by the controller computer to the solenoid valves. The controller computer is tasked with performing real-time planning and computation of control commands based on the status information, with a communication interval of approximately 50ms (20Hz). The commands are then amplified through an amplification circuit and transmitted to the proportional solenoid valves, which regulate the valve spool motion to distribute flow and generate actuator displacement. 
	
	\section{Excavator Dynamic Model}\label{model}
	
	The subsystem of basic hydraulic manipulator model can be divided into: 1) mechanical arm - hydraulic cylinder, 2) hydraulic cylinder - servo valve, and 3) servo valve - electronic controller.
	
	\subsection{Mechanical Arm - Hydraulic Cylinder}
	Using Lagrangian method, the dynamics of n-link rigid hydraulic manipulator can be described by second-order nonlinear differential equations:
	\begin{equation}\label{eq1}
		{{M}_{q}}(q)\ddot{q}+{{C}_{q}}(q,\dot{q})\dot{q}+{{G}_{q}}(q)=\tau -{{\tau }_{e}}
	\end{equation}
	where $q,\dot{q},\ddot{q}\in {{R}^{n}}$ are the vectors of joint angular position, velocity and acceleration respectively, ${{M}_{q}}(q)\in {{R}^{n\times n}}$ denotes the robot manipulator inertia matrix, ${{C}_{q}}(q,\dot{q})\dot{q}\in {{R}^{n}}$ represents the coriolis and centrifugal torques, ${{G}_{q}}(q)\in {{R}^{n}}$ is the vector of gravitational torques, and $\tau ,{{\tau }_{e}}\in {{R}^{n}}$ represent the vector of joint torques originating from the hydraulic actuators, and lumped external force disturbances.
	
	The transformation between the joint space and the actuator space can be described as follows:
	\begin{equation}\label{eq2}
		\begin{aligned}
			& q=\varphi (y) \\ 
			& \dot{q}={{J}^{-1}}(q)\dot{y} \\ 
			& \ddot{q}={{J}^{-1}}(q)\ddot{y}-{{J}^{-1}}(q)\dot{J}(q)\dot{q} \\ 
		\end{aligned}
	\end{equation}
	where $y,\dot{y},\ddot{y}\in {{R}^{n}}$ represent the piston displacement, velocity and acceleration corresponding to mechanical arms. $\varphi$ is the kinematic transition between piston displacement and joint angle. $J,{{J}^{-1}}\in {{R}^{n\times n}}$ are jacobian matrix and its inverse matrix respectively. The dynamics model can be transformed from the joint space to the actuator space:
	\begin{equation}\label{eq3}
		{{M}_{y}}(y)\ddot{y}+{{C}_{y}}(y,\dot{y})\dot{y}+{{G}_{y}}(y)=F-{{F}_{e}}
	\end{equation}
	where 
	\begin{equation}\label{eq4}
		\begin{aligned}
			& {{M}_{y}}(y)={{{J}^{-1}}^{T}}{{M}_{q}}{{J}^{-1}} \\ 
			& {{C}_{y}}(y,\dot{y})={{{J}^{-1}}^{T}}[{{C}_{q}}-{{M}_{q}}{{J}^{-1}}\dot{J}]{{J}^{-1}} \\ 
			& {{G}_{y}}(y)={{{J}^{-1}}^{T}}{{G}_{q}} \\ 
			& F={{{J}^{-1}}^{T}}\tau, {{F}_{e}}={{{J}^{-1}}^{T}}{{\tau }_{e}}\\ 
		\end{aligned}
	\end{equation}
	are the transformation matrix of the corresponding joint space. ${{M}_{q}}(q)$, ${{M}_{y}}(y)$ are both positive definite symmetric matrices, and (${{M}_{q}}(q)-2{{C}_{q}}(q,\dot{q})$), (${{M}_{y}}(y)-2{{C}_{y}}(y,\dot{y})$) are skew-symmetric matrices.
	
	\subsection{Hydraulic Cylinder - Servo Valve}
	The single rod hydraulic actuator dynamic model involving the friction effect can be expressed as: 
	\begin{equation}\label{eq5}
		F={{P}_{1}}{{A}_{1}}-{{P}_{2}}{{A}_{2}}-{{F}_{f}}
	\end{equation}
	where ${P}_{1}$ and ${P}_{2}$ are the forward and return pressures of the cylinder, ${A}_{1}$ and ${A}_{2}$ are the piston areas facing the extend chamber and the retract chamber. ${F}_{f}$ represents the ideal friction force, ${{F}_{f}}={{F}_{v}}\dot{y}$.
	
	Neglecting external leakage, the differential equations for the cylinder pressures ${P}_{1}$ and ${P}_{2}$ can be expressed as:
	\begin{equation}\label{eq6}
		\begin{aligned}
			& {{{\dot{P}}}_{1}}=\frac{{{\beta }_{e}}}{{{V}_{1}}}[{{Q}_{1}}-{{A}_{1}}\dot{y}-{{C}_{t}}({{P}_{1}}-{{P}_{2}})] \\ 
			& {{{\dot{P}}}_{2}}=\frac{{{\beta }_{e}}}{{{V}_{2}}}[{{A}_{2}}\dot{y}+{{C}_{t}}({{P}_{1}}-{{P}_{2}})-{{Q}_{2}}] \\ 
		\end{aligned}
	\end{equation}
	where ${\beta }_{e}$ is the effective bulk modulus of the hydraulic fluid, ${{V}_{1}}={{V}_{01}}+{{A}_{1}}y$, ${{V}_{2}}={{V}_{02}}-{{A}_{2}}y$ are the volumes of the extend chamber and the retract chamber, and ${V}_{01}$, ${V}_{02}$ are the volumes of the two chambers when the piston is in the middle. ${Q}_{1}$ and ${Q}_{2}$ represent the supply flow rate to the forward chamber and the return flow rate to the return chamber respectively. ${C}_{t}$ is the coefficient of the internal leakage of the cylinder.
	
	The dynamics of the piston is achieved by regulating the flow to the chamber of the hydraulic cylinder through the servo valve, and the flow control relationship can be given as:
	\begin{equation}\label{eq7}
		\begin{aligned}
			& {{Q}_{1}}=\frac{{{C}_{d1}}{{\omega }_{1}}}{\sqrt{\rho }}{{y}_{v}}{{\varphi }_{1}}({{P}_{1}},\text{sign}({{y}_{v}})) \\ 
			& {{Q}_{2}}=\frac{{{C}_{d2}}{{\omega }_{2}}}{\sqrt{\rho }}{{y}_{v}}{{\varphi }_{2}}({{P}_{2}},\text{sign}({{y}_{v}})) \\ 
		\end{aligned}
	\end{equation}
	where 
	\begin{equation}\label{eq8}
		\begin{aligned}
			& {{\varphi }_{1}}({{P}_{1}},\text{sign}({{y}_{v}}))=\left\{ \begin{aligned}
				& \begin{matrix}
					\sqrt{{{P}_{s}}-{{P}_{1}}} & {{y}_{v}}\ge 0  \\
				\end{matrix} \\ 
				& \begin{matrix}
					\sqrt{{{P}_{1}}-{{P}_{0}}} & {{y}_{v}}<0  \\
				\end{matrix} \\ 
			\end{aligned} \right. \\ 
			& {{\varphi }_{2}}({{P}_{2}},\text{sign}({{y}_{v}}))=\left\{ \begin{aligned}
				& \begin{matrix}
					\sqrt{{{P}_{2}}-{{P}_{0}}} & {{y}_{v}}\ge 0  \\
				\end{matrix} \\ 
				& \begin{matrix}
					\sqrt{{{P}_{s}}-{{P}_{2}}} & {{y}_{v}}<0  \\
				\end{matrix} \\ 
			\end{aligned} \right. \\ 
		\end{aligned}
	\end{equation}
	and ${C}_{d}$ is the orifice flow coefficient, ${\omega }$ is the area gradient of the servo valve spool, and ${\rho }$ is the density of hydraulic oil. ${y}_{v}$ is the servo valve position. ${P}_{s}$ and ${P}_{0}$ are the supply pump pressure and return pressure respectively.
	
	\subsection{Servo Valve - Electronic Components}
	The dynamics of the servo valve can be expressed as:
	\begin{equation}\label{eq9}
		{{T}_{v}}{{\dot{y}}_{v}}=-{{y}_{v}}+{{k}_{v}}\sigma (u)
	\end{equation}
	where ${T}_{v}$ is the time constant, and ${k}_{v}$ is the servo valve gain. The time constant of the servo valve can be neglected because it is much smaller than the rest of the hydraulic circuits \cite{b19}. Then, the servo valve dynamics can be simplified as ${{y}_{v}}={{k}_{v}}\sigma (u)$. ${u}$ is the input signal, and the $\sigma (u)$ is the nonlinearity in signal transmission process.
	
	In this work, we consider two typical nonlinearities in hydraulic systems: 1) dead-zone and 2) hysteresis \cite{b20}, as shown in Fig.\ref{deadzone}. The mathematical models of the two nonlinearities are as follows: 
	\begin{figure}[htbp]
		\centerline{\includegraphics[width=\columnwidth]{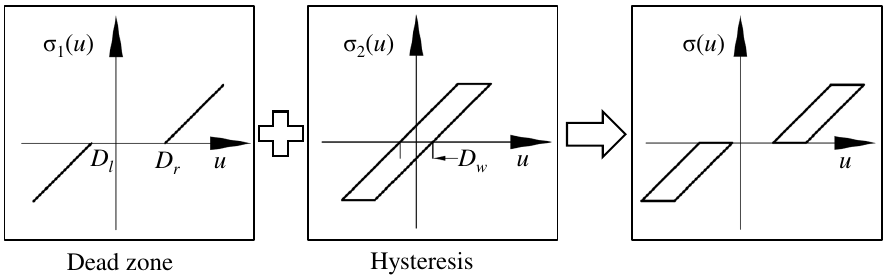}}
		\caption{The nonlinearity composed of dead-zone and hysteresis.}
		\label{deadzone}
	\end{figure}
	
	\begin{equation}\label{eq10}
		\begin{aligned}
			& {{\sigma }_{1}}(u(t))=\left\{ \begin{aligned}
				& \begin{matrix}
					0 & {{D}_{l}}\le u(t)\le {{D}_{r}}  \\
				\end{matrix} \\ 
				& \begin{matrix}
					\begin{aligned}
						& u(t)-{{D}_{r}} \\ 
						& u(t)-{{D}_{l}} \\ 
					\end{aligned} & \begin{aligned}
						& u(t)>{{D}_{r}} \\ 
						& u(t)<{{D}_{l}} \\ 
					\end{aligned}  \\
				\end{matrix} \\ 
			\end{aligned} \right. \\ 
			& {{\sigma }_{2}}(u(t))=\left\{ \begin{aligned}
				& \begin{matrix}
					0 & -{{D}_{w}}/2<u(t)<{{D}_{w}}/2  \\
				\end{matrix} \\ 
				& \begin{matrix}
					\delta [u(t),\dot{u}(t)] & \delta \dot{u}(t)>0  \\
				\end{matrix} \\ 
				& \begin{matrix}
					\delta [u(t),\dot{u}(t)] & \delta \dot{u}(t)<0,\left| u(t)-u({{t}_{0}}) \right|>{{D}_{w}}  \\
				\end{matrix} \\ 
				& \begin{matrix}
					u({{t}_{0}}) & \delta \dot{u}(t)<0,\left| u(t)-u({{t}_{0}}) \right|\le {{D}_{w}}  \\
				\end{matrix} \\ 
			\end{aligned} \right. \\ 
			& \sigma (u(t))={{\sigma }_{2}}({{\sigma }_{1}}(u(t))) \\ 
		\end{aligned}
	\end{equation}
	where $\delta (u,\dot{u})=u-{{D}_{w}}\text{sign(}\dot{u}\text{)}/2$, $D_r, D_l$, and $D_w$ are the width of left dead-zone, right dead-zone and hysteresis, respectively. $u({{t}_{0}})$ is the value when the signal direction changes.
	
	\subsection{Complete dynamic model}
	Assume that all of the hydraulic actuators driving the manipulator links have the similar structure. The state variables are defined as ${{x}_{1}}=q,{{x}_{2}}=\dot{q},{{x}_{3}}=\ddot{q}$ . The dynamic model of the hydraulic manipulator can be formulated in state space: 
	\begin{equation}\label{eq11}
		\left\{ \begin{aligned}
			& {{{\dot{x}}}_{1}}={{x}_{2}} \\ 
			& {{{\dot{x}}}_{2}}={{x}_{3}} \\ 
			& {{{\dot{x}}}_{3}}={{J}^{-1}}{{M}_{x}}^{-1}[{{\kappa }_{1}}-{{\kappa }_{2}}-{{\kappa }_{3}}-{{{\dot{G}}}_{x}}-{{{\dot{F}}}_{e}}]\\
			&\ \ \ \ \ \ -2{{J}^{-1}}\dot{J}{{x}_{3}}-{{J}^{-1}}\ddot{J}{{x}_{2}} \\ 
		\end{aligned} \right.
	\end{equation}
	where 
	\[
	\begin{aligned}
		& {{\kappa }_{1}}=\frac{{{\beta }_{e}}{{k}_{v}}}{\sqrt{\rho }}\left( {{C}_{d1}}{{\omega }_{1}}{{A}_{1}}{{V}_{1}}^{-1}{{\varphi }_{1}}+{{C}_{d2}}{{\omega }_{2}}{{A}_{2}}{{V}_{1}}^{-1}{{\varphi }_{2}} \right)\sigma (u) \\ 
		&\ \ \ \ \ \ -{{\beta }_{e}}J\left( {{A}_{1}}^{2}{{V}_{1}}^{-1}+{{A}_{2}}^{2}{{V}_{1}}^{-1} \right){{x}_{2}}\\
		&\ \ \ \ \ \ -{{\beta }_{e}}{{C}_{t}}\left( {{A}_{1}}{{V}_{1}}^{-1}+{{A}_{2}}{{V}_{1}}^{-1} \right)({{P}_{1}}-{{P}_{2}})-{{{\dot{F}}}_{f}} \\ 
		& {{\kappa }_{2}}=\left( {{{\dot{M}}}_{y}}+{{C}_{y}} \right)\left( \dot{J}{{x}_{2}}+J{{x}_{3}} \right) \\ 
		& {{\kappa }_{3}}={{{\dot{C}}}_{x}}J{{x}_{2}}. \\ 
	\end{aligned}
	\]
	
	Utilizing nonlinear affine transformations, we can describe the dynamic model mentioned above as succinctly as:
	\begin{equation}\label{eq12}
		\left\{ \begin{aligned}
			& {{{\dot{x}}}_{1}}={{x}_{2}} \\ 
			& {{{\dot{x}}}_{2}}={{x}_{3}} \\ 
			& {{{\dot{x}}}_{3}}=f({{x}_{1}},{{x}_{2}},{{x}_{3}})+g({{x}_{1}},{{x}_{2}},{{x}_{3}})u. \\ 
		\end{aligned} \right.
	\end{equation}
	
	\section{Designing Hybrid Control Framework}\label{Mehtod}
	
	This section introduces RevNM and the hybrid control framework for hydraulic manipulators. The third-order nonlinear system is reformulated into a reversible model using reversible transformations. Then, the corresponding model inversion controller is designed with the backstepping method, and the stability is proved by Lyapunov theory. We combine the PD controller and the model inversion controller to form a hybrid control framework. At the end, the neural network framework and training process are explained.
	
	\subsection{RevNM Plant Model}
	First, consider a general third-order nonlinear system:
	\begin{equation}\label{eq13}
		\left\{ \begin{aligned}
			& {{{\dot{x}}}_{1}}={{f}_{1}}({{x}_{1}},{{x}_{2}},{{x}_{3}}) \\ 
			& {{{\dot{x}}}_{2}}={{f}_{2}}({{x}_{1}},{{x}_{2}},{{x}_{3}}) \\ 
			& {{{\dot{x}}}_{3}}={{f}_{3}}({{x}_{1}},{{x}_{2}},{{x}_{3}})+{{g}}({{x}_{1}},{{x}_{2}},{{x}_{3}})u. \\ 
		\end{aligned} \right.
	\end{equation}
	
	A bidirectional mapping relationship between input signals and state variables typically necessitates the invertibility of function ${g}$. By modifying ${g}$ as ${e}^{{S}}$ and incorporating the nonlinear function $T_{i}$ \cite{b21},\cite{b22}, the following forward differential equation is derived:
	\begin{equation}\label{eq14}
		\left\{ \begin{aligned}
			& {{{\dot{x}}}_{1}}={{T}_{1}}(h)+{{x}_{2}} \\ 
			& {{{\dot{x}}}_{2}}={{T}_{2}}(h)+{{x}_{3}} \\ 
			& {{{\dot{x}}}_{3}}={{T}_{3}}(h)+{{e}^{S(h)}}u \\ 
		\end{aligned} \right.
	\end{equation}
	and the corresponding inversion model is:
	\begin{equation}\label{eq15}
		\left\{ \begin{aligned}
			& {{x}_{2}}={{{\dot{x}}}_{1}}-{{T}_{1}}(h) \\ 
			& {{x}_{3}}={{{\dot{x}}}_{2}}-{{T}_{2}}(h) \\ 
			& u={{e}^{-S(h)}}({{{\dot{x}}}_{3}}-{{T}_{3}}(h)) \\ 
		\end{aligned} \right.
	\end{equation}
	where ${{T}_{i}}(h)=w_{i}^{1b}\cdot \phi (w_{i}^{1a}h+b_{i}^{1}), {S}(h)=clamp\cdot \text{atan}(w_{i}^{2b}\cdot \phi (w_{i}^{2a}h+b_{i}^{2}))$ are both multilayer perceptrons (MLP), which are used to approximate nonlinear dynamics. ${h}$ is the set of state parameters, and the $clamp$ is soft clamping for the multiplicative component. $\phi$, $w$, $b$ are the activation function, weight coefficient and bias coefficient respectively. $w$ and $b$ are bounded, $w\le \left| {{w}_{\max }} \right|,b\le \left| {{b}_{\max }} \right|$. 
	
	RevNM enables bidirectional mapping of input and output, allowing for the acquisition of both forward and inverse models in a single training process \cite{b22},\cite{b23}. Compared to constructing a reversible neural network directly, the introducing of physical model information effectively sidesteps the constraint that input and output dimensions must be matched \cite{b18},\cite{b24}, and enhances interpretability.
	
	\begin{figure}[htbp]
		\centerline{\includegraphics[width=\columnwidth]{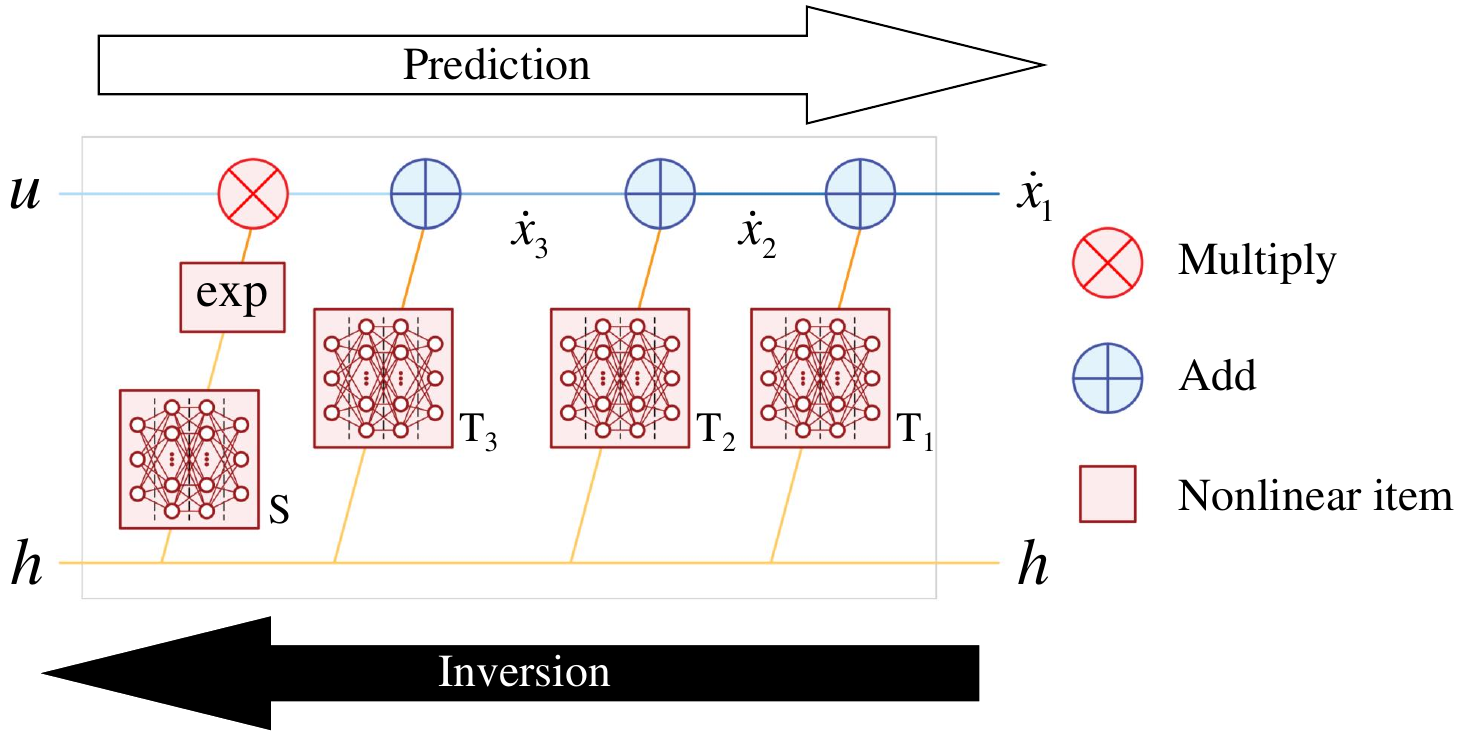}}
		\caption{The framework of RevNM. $h$ encompass a multitude of current available state parameters, such as $[\Delta t, {{x}_{1}}(t),{{x}_{1}}(t-\Delta t),\ldots ,{{x}_{2}}(t),{{x}_{2}}(t-\Delta t),\ldots ,{{x}_{3}}(t),{{x}_{3}}(t-\Delta t),\ldots ]$, determined by the system's state variables.}
		\label{revnm}
	\end{figure}
	
	The basic framework of RevNM is depicted in Fig.\ref{revnm}. The prediction model and inversion model can be implemented through the (\ref{eq14}) and (\ref{eq15}). The prediction model utilizes the current state parameters $h=[{{x}_{1}},{{x}_{2}},{{x}_{3}}]^T$ and input $u$ to calculate $[{{\dot{x}}_{1}},{{\dot{x}}_{2}},{{\dot{x}}_{3}}]^T$ through the forward process. If a stable step $\Delta t$ exists, we can obtain $[\Delta {{\hat{x}}_{1}},\Delta {{\hat{x}}_{2}},\Delta {{\hat{x}}_{3}}]^T$. Meanwhile the inversion model requires desired $[{{\dot{x}}_{1}},{{\dot{x}}_{2}},{{\dot{x}}_{3}}]^T$ the theoretical input signal $\hat{u}$ is then calculated through the backward process.
	
	\subsection{Hybrid Control Framework}
\begin{figure}[htbp]
		\centerline{\includegraphics[width=\columnwidth]{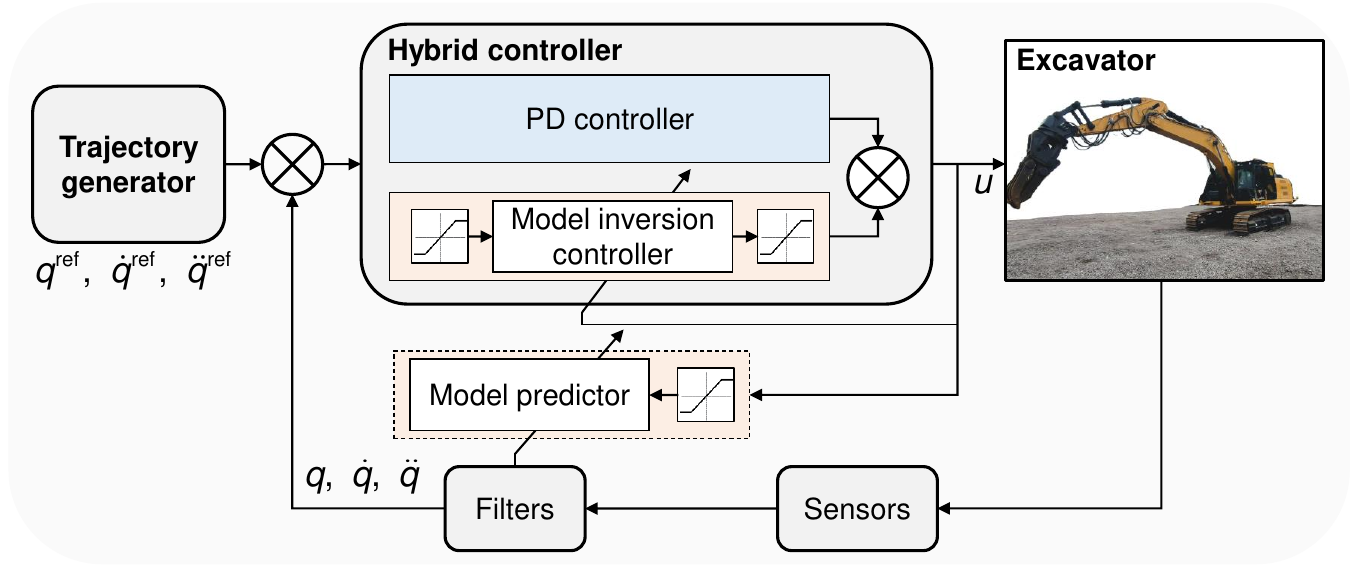}}
		\caption{Hybrid control framework for hydraulic manipulator.}
		\label{hybridcontroller}
	\end{figure}

	The hybrid control framework for RevNM is shown in Fig.\ref{hybridcontroller}, which consists of PD controller and model inversion controller. PD controller is utilized for rapid convergence when the tracking error is large, while also enhancing the robustness. Model inversion controller can provide more precise compensation for the nonlinear dynamics. To alleviate the unknown impacts caused by the fitting errors of neural networks, limiting is necessary for the input and output data of the model inversion controller. Additionally, this work only employs the prediction model for observation, but there is significant potential for its use such as model predictive control (MPC).
	
	\subsection{Model Inversion Controller Design}
	For the RevNM depicted in (\ref{eq14}), define the reference trajectory as $x_{1}^{\text{ref}}$, and $x_{2}^{\text{ref}}$, $x_{3}^{\text{ref}}$ can be obtained through differentiation. Both $x_{1}^{\text{ref}}$, $x_{2}^{\text{ref}}$, and $x_{3}^{\text{ref}}$ are continuous and bounded considering the mechanical structure and limitations in practice. The detailed design process is as follows:
	
	\textit{step 1 :} Let ${{z}_{1}}={{x}_{1}}-x_{1}^{\text{ref}}$, and the derivative of ${z}_{1}$ is:
	\begin{equation}\label{eq16}
		{{\dot{z}}_{1}}={\dot{x}}_{1}-{\dot{x}}_{1}^{\text{ref}}={{T}_{1}}+{{x}_{2}}-x_{2}^{\text{ref}}.
	\end{equation}
	Design a virtual control law:
	\begin{equation}\label{eq17}
		{{\alpha }_{2}}=x_{2}^{\text{ref}}-{{T}_{1}}-{{k}_{1}}{{z}_{1}}
	\end{equation}
	where ${{k}_{1}}>0$ is a positive parameter to be designed.
	
	\textit{step 2 :} Let ${{z}_{2}}={{x}_{2}}-{{\alpha }_{2}}$. Differentiating ${{z}_{2}}$, one can obtain that:
	\begin{equation}\label{eq18}
		{{{\dot{z}}}_{2}}={\dot{x}_{2}}-{{\dot{\alpha }_{2}}}={{T}_{2}}+{{x}_{3}}+{{{\dot{T}}}_{1}}+{{k}_{1}}{{{\dot{z}}}_{1}}-x_{3}^{\text{ref}}. \\ 
	\end{equation}
	We can design the virtual control law ${{\alpha }_{3}}$ as follows:
	\begin{equation}\label{eq19}
		{{\alpha }_{3}}=x_{3}^{\text{ref}}-{{T}_{2}}-{{\dot{T}}_{1}}-{{z}_{1}}-{{k}_{1}}{{\dot{z}}_{1}}-{{k}_{2}}{{z}_{2}}
	\end{equation}
	where ${{k}_{2}}>0$ is a positive parameter to be designed.
	
	\textit{step 3 :} Let ${{z}_{3}}={{x}_{3}}-{{\alpha }_{3}}$. Based on (\ref{eq14}), the time derivative of ${{z}_{3}}$ can be obtained that:
	\begin{equation}\label{eq20}
		{{\dot{z}}_{3}}={{T}_{3}}+{{e}^{{{S}_{3}}}}u+({{\dot{z}}_{1}}+{{\dot{T}}_{2}}+{{\ddot{T}}_{1}}+{{k}_{1}}{{\ddot{z}}_{1}}+{{k}_{2}}{{\dot{z}}_{2}}-\dot{x}_{3}^{\text{ref}}).
	\end{equation}
	Let ${{k}_{3}}>0$, the control law can be described as:
	\begin{equation}\label{eq21}
		u={{e}^{-S}}(\dot{x}_{3}^{\text{ref}}-{{T}_{3}}-{{\dot{T}}_{2}}-{{\ddot{T}}_{1}}-{{R}_{1}}{{z}_{1}}-{{R}_{2}}{{\dot{z}}_{1}}-{{R}_{3}}{{\ddot{z}}_{1}})
	\end{equation}
	where 
	\[
	\begin{aligned}
		& {{R}_{1}}={{k}_{1}}{{k}_{2}}{{k}_{3}}+{{k}_{1}}+{{k}_{3}} \\ 
		& {{R}_{2}}={{k}_{1}}{{k}_{2}}+{{k}_{1}}{{k}_{3}}+{{k}_{2}}{{k}_{3}}+2 \\ 
		& {{R}_{3}}={{k}_{1}}+{{k}_{2}}+{{k}_{3}}. \\ 
	\end{aligned}
	\]
	
	\subsection{Stability Analysis}
	First of all, let:
	\begin{equation}\label{eq22}
		\left\{ \begin{aligned}
			& {{z}_{1}}={{x}_{1}}-x_{1}^{\text{ref}} \\ 
			& {{z}_{2}}={{x}_{2}}-{{\alpha }_{2}} \\ 
			& {{z}_{3}}={{x}_{3}}-{{\alpha }_{3}}. \\ 
		\end{aligned} \right.
	\end{equation}
	
	The dynamics with account for fitting error and disturbances can be obtained as:
	\begin{equation}\label{eq23}
		\left\{ \begin{aligned}
			& {{{\dot{z}}}_{1}}={\dot{x}_{1}}-\dot{x}_{1}^{\text{ref}}+{{d}_{1}} \\ 
			& {{{\dot{z}}}_{2}}={\dot{x}_{2}}-{\dot{\alpha }_{2}}+{{d}_{2}} \\ 
			& {{{\dot{z}}}_{3}}={\dot{x}_{3}}-{\dot{\alpha }_{3}}+{{d}_{3}}. \\ 
		\end{aligned} \right.
	\end{equation}
	
	\textit{Assumption 1 :} The unknown disturbances ${{d}_{i}}$ caused by fitting errors and other external influences are bounded:$\left\| {{d}_{i}} \right\|\le \left\| {{d}_{\max }} \right\|$, ${{d}_{\max }}$ is upper bound of ${d}_{i}$. The assumption is valid due to reliable learning performances \cite{b7} and finite physical energy in practice.
	
	\textit{Theorem 1 :} Based on \textit{Assumption 1}, consider the third-order nonlinear system constructed by (\ref{eq23}), under the controller (\ref{eq21}), all signals are uniformly ultimately bounded.
	
	\textit{Proof :} For \textit{step 1}, we design the Lyapunov function candidate as:
	\begin{equation}\label{eq24}
		{{V}_{1}}=\frac{1}{2}{{z}_{1}}^{2}.
	\end{equation}
	Differentiating (\ref{eq24}) and invoking (\ref{eq17}), one can obtain that:
	\begin{equation}\label{eq25}
		{{\dot{V}}_{1}}=-{{k}_{1}}z_{1}^{2}+{{z}_{1}}{{z}_{2}}+{{d}_{1}}{{z}_{1}}.
	\end{equation}
	According to Young’s inequality, it is easy to obtain:
	\begin{equation}\label{eq26}
		{{\dot{V}}_{1}}\le -({{k}_{1}}-1){{\left\| {{z}_{1}} \right\|}^{2}}+\frac{1}{2}{{\left\| {{z}_{2}} \right\|}^{2}}+\frac{1}{2}{{\left\| {{d}_{1}} \right\|}^{2}}.
	\end{equation}
	
	Then, the Lyapunov function candidate \textit{step 2} can be designed as:
	\begin{equation}\label{eq27}
		{{V}_{2}}={{V}_{1}}+\frac{1}{2}{{z}_{2}}^{2}.
	\end{equation}
	
	With Young’s inequality and (\ref{eq19}), the time derivative of (\ref{eq27}) is obtained as:
	\begin{equation}\label{eq28}
		\begin{aligned}
			& {{{\dot{V}}}_{2}}={{{\dot{V}}}_{1}}-{{k}_{2}}z_{2}^{2}+{{z}_{2}}{{z}_{3}}+{{d}_{2}}{{z}_{2}} \\ 
			& \ \ \ =-{{k}_{1}}z_{1}^{2}-{{k}_{2}}z_{2}^{2}+{{z}_{2}}{{z}_{3}}+{{d}_{1}}{{z}_{1}}+{{d}_{2}}{{z}_{2}} \\ 
			& \ \ \ \le -({{k}_{1}}-\frac{1}{2}){{\left\| {{z}_{1}} \right\|}^{2}}-({{k}_{2}}-1){{\left\| {{z}_{2}} \right\|}^{2}}+\frac{1}{2}{{\left\| {{z}_{3}} \right\|}^{2}} \\
			& \ \ \ +\frac{1}{2}{{\left\| {{d}_{1}} \right\|}^{2}}+\frac{1}{2}{{\left\| {{d}_{2}} \right\|}^{2}}. \\ 
		\end{aligned}
	\end{equation}
	
	Similarly, for \textit{step 3}, we have
	\begin{equation}\label{eq29}
		{{V}_{3}}={{V}_{2}}+\frac{1}{2}{{z}_{3}}^{2}
	\end{equation}
	and
	\begin{equation}\label{eq30}
		\begin{aligned}
			& \dot{V}=-{{k}_{1}}{{z}_{1}}^{2}-{{k}_{2}}{{z}_{2}}^{2}-{{k}_{3}}{{z}_{3}}^{2}+{{d}_{1}}{{z}_{1}}+{{d}_{2}}{{z}_{2}}+{{d}_{3}}{{z}_{3}} \\ 
			& \le -({{k}_{1}}-\frac{1}{2}){{\left\| {{z}_{1}} \right\|}^{2}}-({{k}_{2}}-\frac{1}{2}){{\left\| {{z}_{2}} \right\|}^{2}}-({{k}_{3}}-\frac{1}{2}){{\left\| {{z}_{3}} \right\|}^{2}} \\
			& \ \ \ +1.5{\left\| {d}_{\max } \right\|}^{2}. \\
		\end{aligned}
	\end{equation}
	
	Let $\gamma = 1.5{{d}_{\max }}^{2}$. Selecting ${{k}_{1}}>0.5,{{k}_{2}}>0.5,{{k}_{3}}>0.5$, $\dot{V}$ can be written as:
	\begin{equation}\label{eq31}
		\dot{V}\le -\lambda V+\gamma
	\end{equation}
	where $\lambda =\min [{{k}_{1}}-0.5,{{k}_{2}}-0.5,{{k}_{3}}-0.5]$.
	
	According to the definition of $V(t)$ and the boundedness theorem, the following can be deduced: 
	\begin{equation}\label{eq32}
		0\le V(t)\le V(0){{e}^{-\lambda t}}+\frac{\gamma }{\lambda }(1-{{e}^{-\lambda t}}).
	\end{equation}
	When $t\to \infty $, it can be inferred that $V\to \gamma /\lambda $, and all error signals in (\ref{eq23}) are UUB \cite{b25}.  The proof is completed.
	
	\subsection{Training}
	Python and the PyTorch library are utilized for programming and training. Each sub-network, $T_i$ and $S$, is composed of two hidden layers with a width of 64 nodes, which are composed of nodes exhibiting gradients, and employ the Relu activation function.
	
	To avoid potential training issues arising from the wide range and large magnitude of physical parameter values, pre-processing of all sampling data is necessary. Due to  the presence of sensor noise, filtering is required after acquiring the motion data. Datasets undergo min-max normalization to standardize within the range of (0, 1). During each training iteration, a certain batch of input and output data, will be drawn from the sampling data set. Weights and gradients are updated using the adams optimizer. Both forward and inverse losses are optimized as presented in \cite{b26}. The effect of bi-directional training is evaluated using mean squared error (MSE) loss between the actual values and the training values computed in both the forward (LossY) and inversion (LossX) directions for each batch. The losses are recorded after each training iteration until they converge. 
	
	\section{Experimental Verification}\label{Experiment}
	
	\subsection{Amesim and Simulink Co-simulation}
	
\begin{figure}[htbp]
		\centerline{\includegraphics[width=\columnwidth]{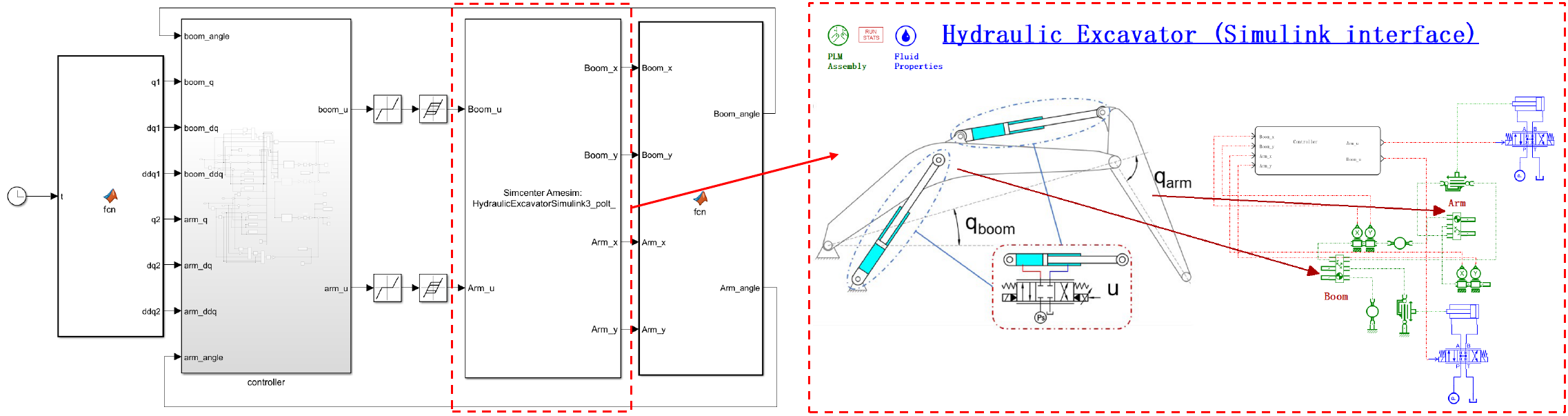}}
		\caption{Amesim and Simulink co-simulation. The simulations employ a fixed-step and ode8 solver. Input signal adopts a zero-order holder, and the sampling time is randomly ranging from 1 to 10 ms.}
		\label{simulation}
	\end{figure}

	First, the RevNM and hybrid control framework are verified through simulation. A hydraulic excavator (Simulink interface) from Amesim demo library is utilized, incorporating the two-degrees-of-freedom for boom and arm. The hybrid control framework is developed with s-function in Simulink, as shown in Fig.\ref{simulation}. The joint angles cannot be directly acquired in Amesim, we calculate them in Simulink based on the coordinates of joint endpoints. The nonlinearities of dead-zone and hysteresis are added during signal transmission in Simulink. The main parameters are shown in the Table \ref{tab1}.
	
	\begin{table}[htbp]
		\centering
		\caption{Co-simulation Parameters}
		\begin{tabular}{p{20pt}p{100pt}p{95pt}}
				\toprule  
				\centering Symbol& 
				Explanation& 
				Value \\ 
				\midrule  
				\centering $P_s$ & pump pressure & 200 bar\\
				\centering $Q_{max}$ & servo valve maximum flow rate & 600 L/min\\
				\centering $D$ & piston diameter & boom: 0.35 m, arm: 0.18 m\\
				\centering $d$ & rod diameter & boom: 0.22 m, arm: 0.125 m\\
				\centering $L$ & length of stroke & boom: 1.8 m, arm: 1.7 m\\
				\centering $a$ & length of link & boom: 7.2 m, arm: 2.9 m\\
				\centering $v_{isc}$ & viscous friction coefficient & 100000 N/(m/s)\\
				\centering $C_t$ & internal leakage coefficient & 0.005 L/min/bar\\
				\centering $M$ & mass & boom: 8000 kg, arm: 2920 kg\\
				\centering $J$ & moment of inertia & boom: 38500 kg·m$^{2}$, arm: 3600 kg·m$^{2}$\\
				\centering $b_z$ & rotary damping coefficient & 10000 N·m/(rev/min)\\
				\centering $[D_l,D_r]$ & dead-zone & boom: [-0.2, 0.1], arm: [-0.1, 0.2]\\
				\centering $D_w$ & hysteresis & position: 0, width: 0.05\\
				\bottomrule  
			\end{tabular}
			\label{tab1}
	\end{table}
	
\begin{figure}[htbp]
		\centerline{\includegraphics[width=\columnwidth]{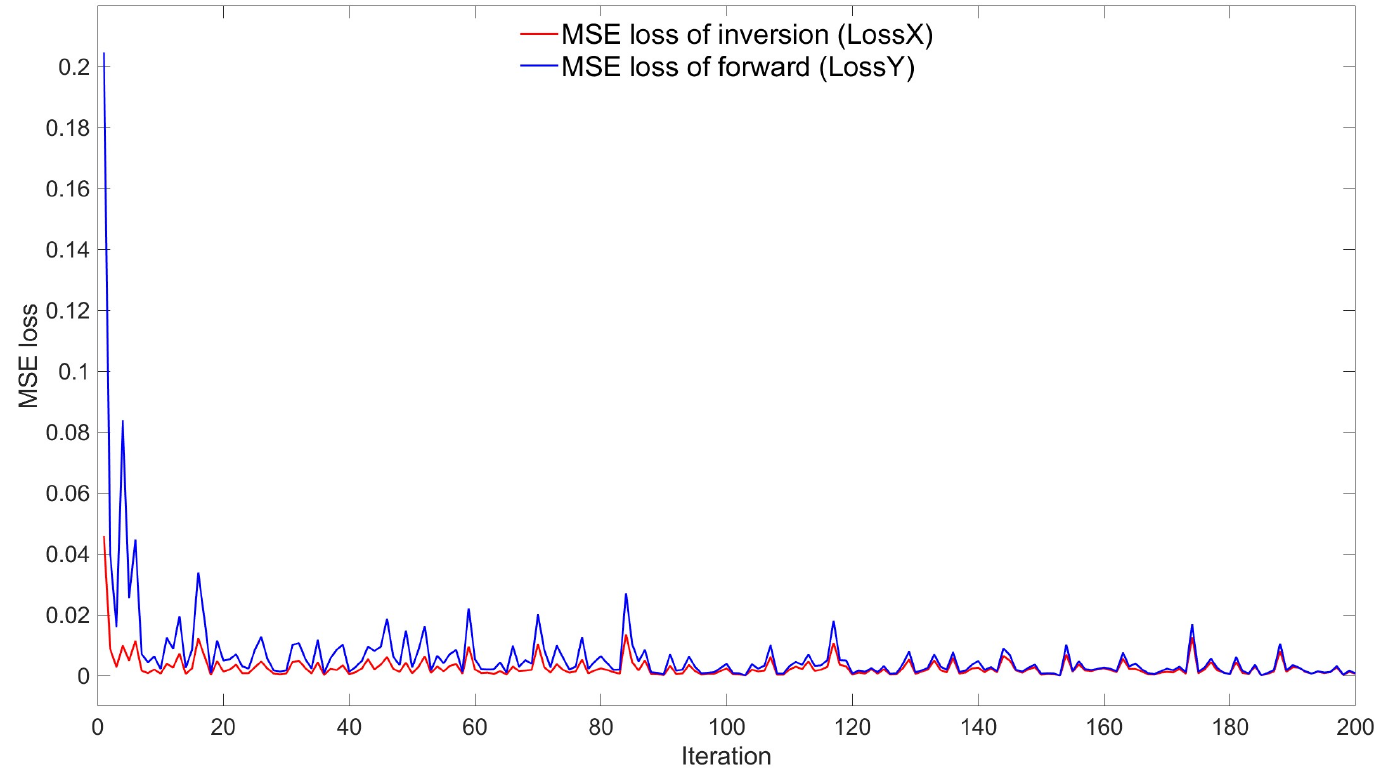}}
		\caption{Convergence process estimated by LossX and LossY. After about 100 iterations, all of the lines become relatively horizontal and smooth, suggesting that the LossX and LossY become convergent.}
		\label{mseloss}
	\end{figure}

	We conduct 10000 simulations by giving random sinusoidal signals, and collect motion data equivalent to 50 hours. Afterward, the offline data are uploaded to the server. The RevNM is trained with Intel I9 12900 3.8 GHz CPU, 64 GB RAM, and NVIDIA GeForce RTX 3090 GPU. The LossX and LossY of the training process are depicted in Fig.\ref{mseloss}, exhibiting a gradual convergence following multiple iterations.
	
	Desired trajectory is set in cartesian space as $x=6.80+\cos (0.2\pi t)$, $y=-2.12+\sin (0.2\pi t)$. The initial position of the manipulator end is set to $x = 7.80m$, $y = -2.12m$.
	
	\begin{figure}[htbp]
		\centerline{\includegraphics[width=\columnwidth]{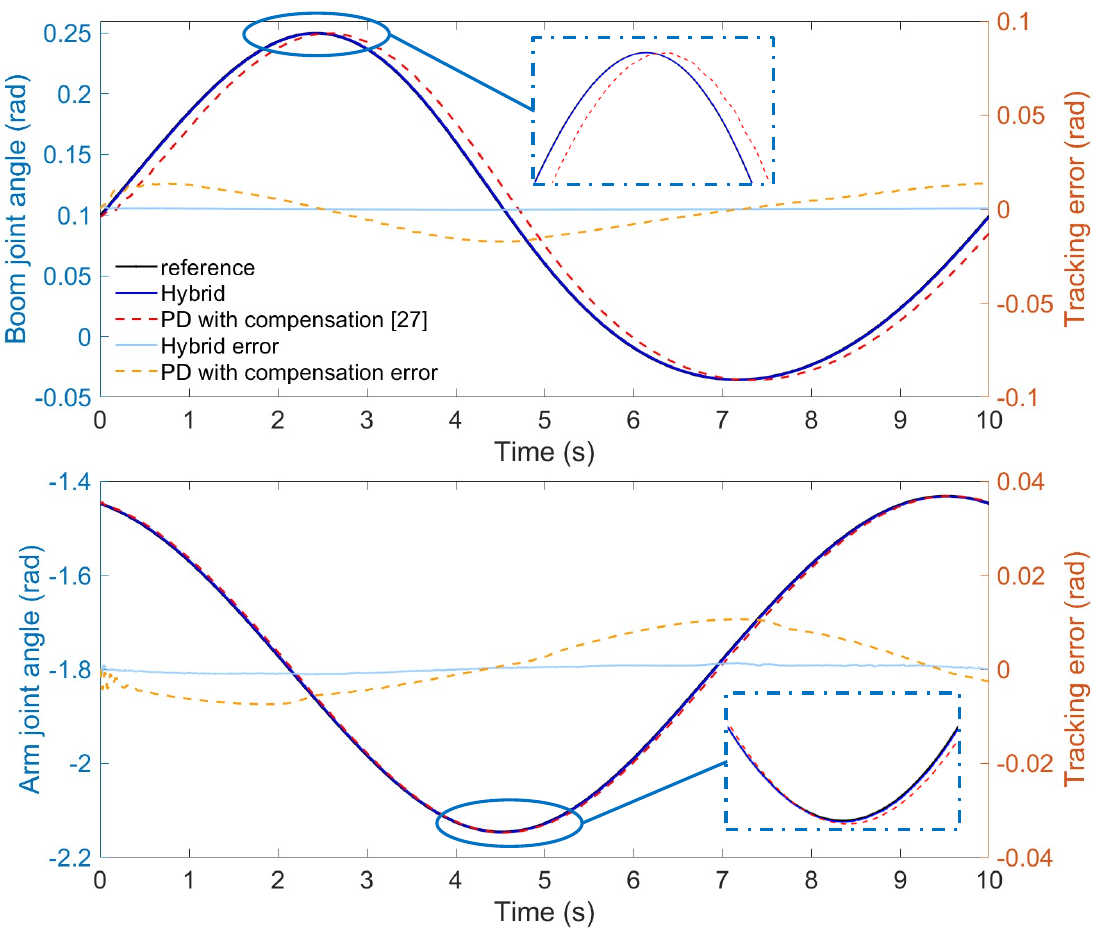}}
		\caption{Trajectory tracking of boom and arm. To facilitate a more discerning comparison of tracking performance, we employ a dual-Y-axis line chart. The position of the tracking curve is plotted on the left Y-axis, while the tracking error is shown on the right Y-axis.}
		\label{simulationjoint}
	\end{figure}
	
\begin{figure}[htbp]
		\centerline{\includegraphics[width=\columnwidth]{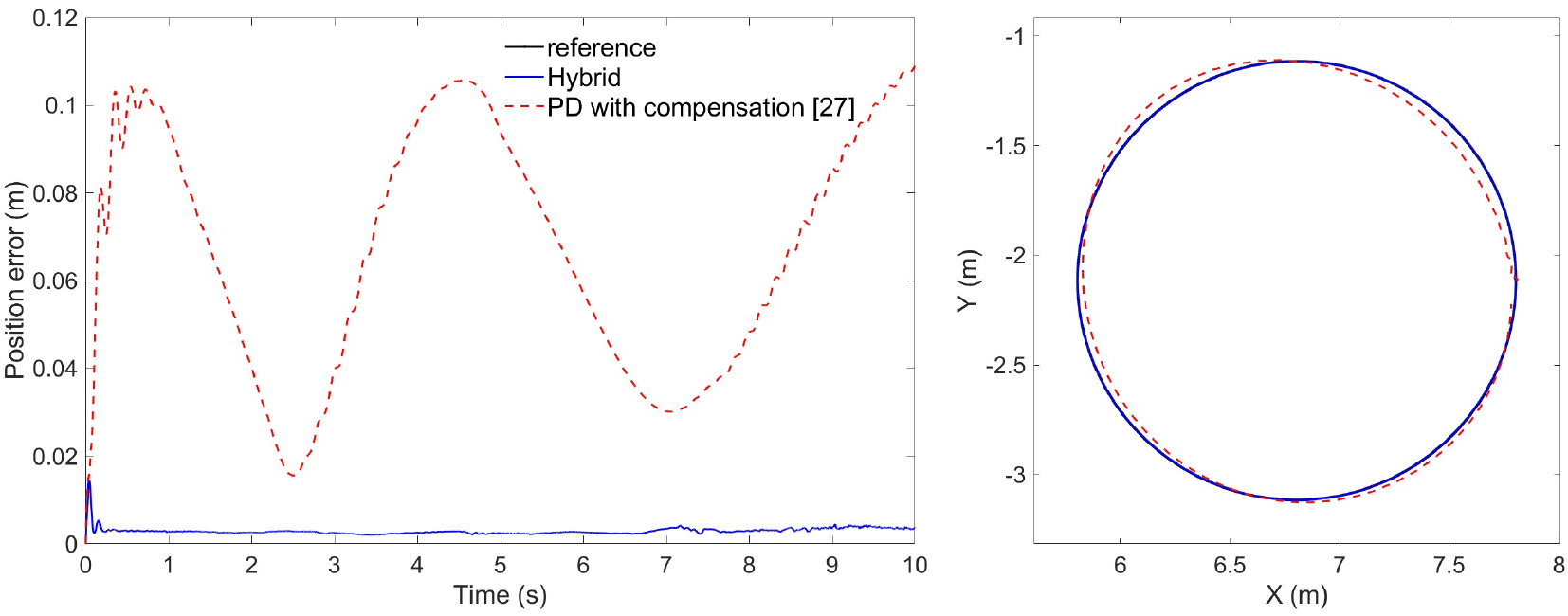}}
		\caption{Trajectory tracking in cartesian space.}
		\label{simulationcircle}
	\end{figure}

	Comparative simulation analysis between the hybrid control framework and PD controller with compensation \cite{b27} has been performed. The PD controller selects the following parameters $k_{P}^{\text{boom}} = 200$, $k_{D}^{\text{boom}} = 0.16$, and $k_{P}^{\text{arm}} = 160$, $k_{D}^{\text{arm}} = 0.2$. Fig.\ref{simulationjoint} shows the tracking comparison of the joint angle under hybrid control framework and PD controller with compensation. It is evident that the proposed framework facilitates convergence of tracking error to a minimal domain. Fig.\ref{simulationcircle} shows the comparison of tracking performance in cartesian space. It can be observed that the proposed framework achieves precise position following.
	
\begin{figure}[htbp]
		\centerline{\includegraphics[width=\columnwidth]{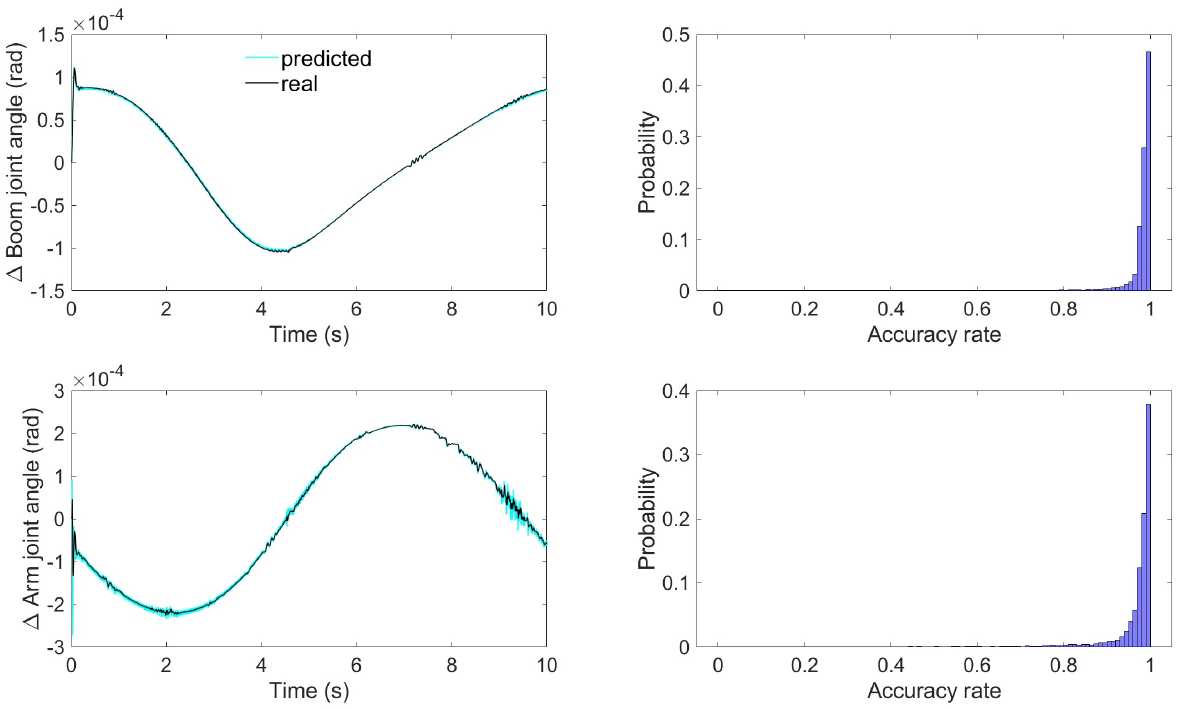}}
		\caption{The predictions of RevNM for $\Delta q$ and the actual $\Delta q$.}
		\label{prediction}
	\end{figure}

	The predictive capability of RevNM has also been validated, as it forecasts the state for a future time period by using the current state and input signal. The results are illustrated in Fig.\ref{prediction}. The deviation between the predicted and actual positions is evaluated using an accuracy metric $\eta =1-\left| \frac{\Delta \hat{x}-\Delta x}{\Delta x} \right|$, where $\Delta x$ and $\Delta \hat{x}$ are actual and predicted change respectively. We categorize the accuracy of each prediction into 1\% intervals on a histogram and compute the expected value to represent the overall trajectory accuracy. The boom and arm prediction model exhibits 0.95 and 0.90.
	
	\subsection{Experiments on Hydraulic Excavator}
	Experiments are conducted on the boom and arm of a commercial 39-ton heavy-load hydraulic excavator, as these two joints have significant impact on the trajectory tracking. However, we believe that the proposed control framework can be extended to the entire excavator by collecting information from more joints. For the offline learning process, motion data are collected by making the robotic arm track randomly given cubic spline curves in the workspace with a manufacturer-provided PD controller. The parameters of the PD controller are set to $k_P = 300$, $k_D = 3$ for the boom and $k_P = 200$, $k_D = 1.8$ for the arm. We collect approximately 0.3 million time steps, which amounts to about 4.2 hours of data. The training process is consistent with the simulation. 
	
	To evaluate the control performance, we select the end position of the arm joint ${{p}_{arm}}=({{p}_{x}},{{p}_{y}})\in {{R}^{2}}$ to trace a circular trajectory, where ${{p}_{x}},{{p}_{y}}\in R$ represent the horizontal and vertical coordinates of the endpoint, respectively. The formulas for circular trajectory is given by: $x=6.77+\cos (0.1\pi t)$, $y=1.39+\sin (0.1\pi t)$. We compute the RMSE for both path following error and trajectory tracking error, and the calculation formulas as shown in (\ref{eq33}).
	\begin{equation}\label{eq33}
		\begin{aligned}
			& \text{RMS}{{\text{E}}_{\text{path}}}=\underset{{{t}_{0}}\le \tau \le {{t}_{f}}}{\mathop{\min }}\,\sqrt{\frac{\sum{{{\left\| {{p}_{t}}-p_{\tau }^{ref} \right\|}_{2}}}}{n}} \\ 
			& \text{RMS}{{\text{E}}_{\text{trajectory}}}=\sqrt{\frac{\sum{{{\left\| {{p}_{t}}-p_{t}^{ref} \right\|}_{2}}}}{n}}. \\ 
		\end{aligned}
	\end{equation}
	
	\begin{figure}[htbp]
		\centerline{\includegraphics[width=\columnwidth]{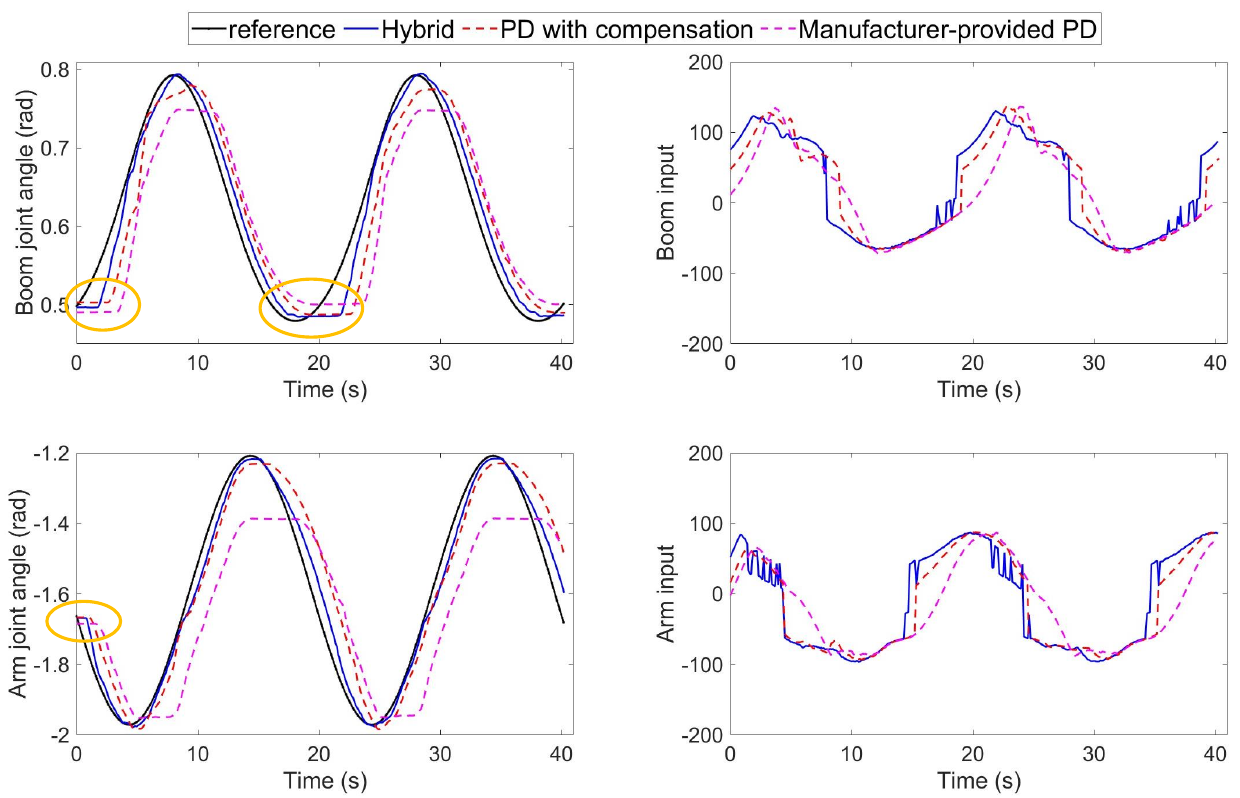}}
		\caption{Trajectory tracking in joint space and input signal. Yellow zone highlights the positions where no flow outputs despite the presence of input signals, due to insufficient output flow caused by changes in the pump swashplate angle. The range of input signal u is [-200, 200].}
		\label{experimentjoint}
	\end{figure}
	
	\begin{figure}[htbp]
		\centerline{\includegraphics[width=\columnwidth]{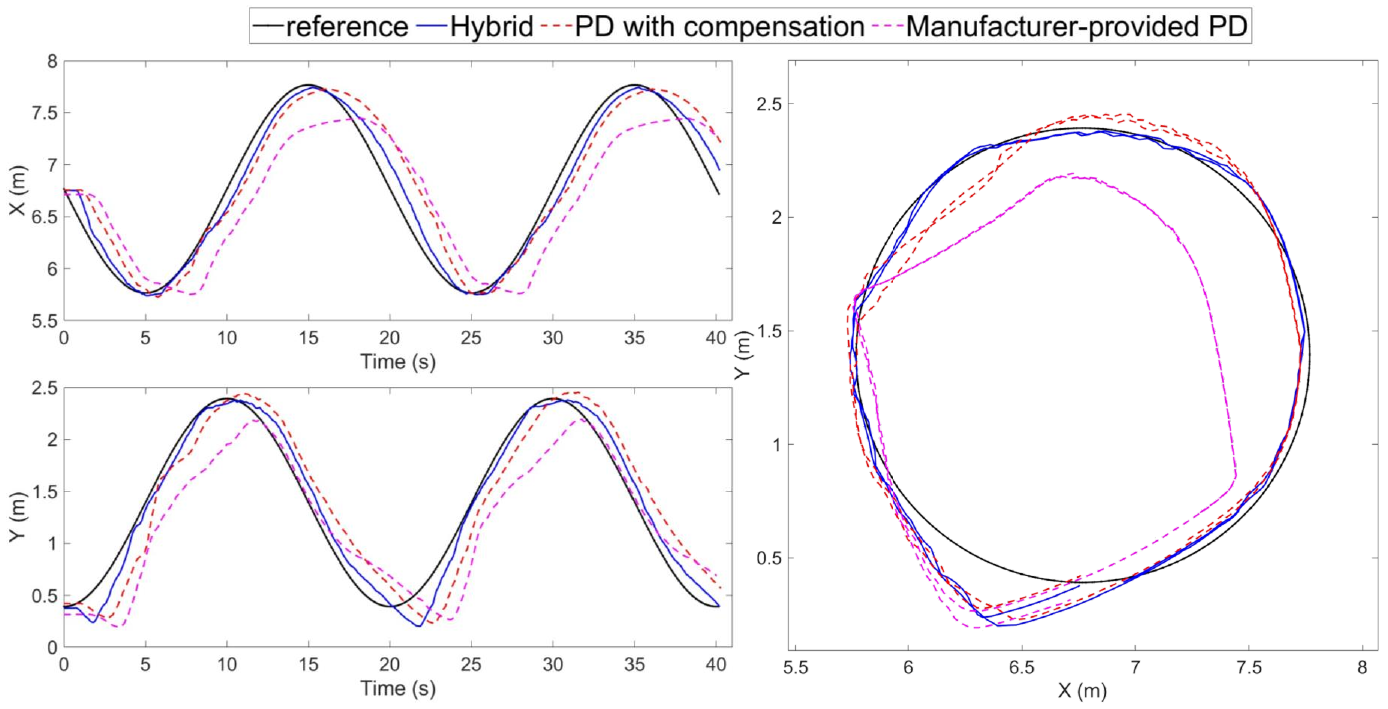}}
		\caption{Trajectory tracking in cartesian space.}
		\label{experimentcircle}
	\end{figure}
	
	\begin{table}[htbp]
		\centering
		\caption{Results of experiment}
		\begin{tabular}{p{35pt}p{35pt}p{65pt}p{80pt}} 
				\toprule  
				\centering RMSE& 
				Hybrid controller& 
				PD controller with compensation \cite{b27}&
				Manufacturer-provided PD controller
				\\ 
				\midrule  
				\centering $path$ & 5.7 cm & 10.6 cm & 27.4 cm\\
				\centering $trajectory$ & 13.5 cm & 28.2 cm & 48.7 cm\\
				\bottomrule  
			\end{tabular}
			\label{tab2}
	\end{table}
	
	The calculation of RMSE for trajectory tracking begins from the moment when boom starts moving, as depicted in Fig.\ref{experimentjoint}. This is due to the excavator's built-in logic whereby the valve spool control commands also affect the inclination angle of the piston pump's swashplate. As comparison, we also conduct experiments on the performance of PD controller with compensation \cite{b27} , and manufacturer-provided PD controller. The results are detailed in Table \ref{tab2} and Fig.\ref{experimentcircle}. The manufacturer-provided PD controller without any compensation struggles to track the trajectory. With dead-zone compensation, the RMSE for path following reaches up to 10.6 cm. Our control framework achieves a path following RMSE of 5.7 cm and a trajectory RMSE of 13.5 cm.
	
	\section{Conclusion}
	To achieve automated operation of industrial heavy-load hydraulic manipulators, this work proposes a data-driven modeling and corresponding hybrid motion control framework to overcome the challenges associated with control performance. We integrate prior physical model information and reversible transformations to develop RevNM, rather than directly constructing mathematical models or simplistic black-box models. The corresponding hybrid control framework consists of model inversion controller and PD controller. We validated the proposed framework through simulation and experimentation, and the control performance is significantly improved. Furthermore, RevNM and the corresponding control framework can be widely applied to different plants based on physical model analysis, and have certain practical value in industry.
	
	Future work includes: 1) Integrating the model predictor with MPC; 2) Designing controllers based on the advanced control algorithms such as sliding mode or non-smooth methods to enhance control performance; 3) Applying the framework to practical engineering tasks.


\begin{thebibliography}{00}
		
		\bibitem{b1} L. Zhang et al., “An autonomous excavator system for material loading tasks,” \emph{Sci. Robot.}, vol. 6, no. 55, Jun, 2021, Art. no. eabc3164, doi: 10.1126/scirobotics.abc3164.
		
		\bibitem{b2} K. Kim, M. Kim, D. Kim, and D. J. Lee, “Modeling and velocity-field control of autonomous excavator with main control valve,” \emph{Automatica}, vol. 104, pp. 67-81, Jun, 2019, doi: 10.1016/j.automatica.2019.02.041.
		
		\bibitem{b3} S. Tafazoli, S. E. Salcudean, K. Hashtrudi-Zaad, and P. D. Lawrence, “Impedance control of a teleoperated excavator,” \emph{IEEE Trans. Control Syst. Technol.}, vol. 10, no. 3, pp. 355-367, May, 2002, doi: 10.1109/87.998021.
		
		\bibitem{b4} W. Kim, D. Won, D. Shin, and C. C. Chung, “Output feedback nonlinear control for electro-hydraulic systems,” \emph{Mechatronics}, vol. 22, no. 6, pp. 766-777, Sep, 2012, doi: 10.1016/j.mechatronics.2012.03.008.
		
		\bibitem{b5} S. Dadhich, U. Bodin, and U. Andersson, “Key challenges in automation of earth-moving machines,” \emph{Autom. Constr.}, vol. 68, pp. 212-222, Aug, 2016, doi: 10.1016/j.autcon.2016.05.009.
		
		\bibitem{b6} K. Hornik, M. Stinchcombe, and H. White, “Multilayer feedforward networks are universal approximators,” \emph{Neural Netw.}, vol. 2, no. 5, pp. 359-366, Jan, 1989, doi: 10.1016/0893-6080(89)90020-8.
		
		\bibitem{b7} D. Yarotsky, “Error bounds for approximations with deep ReLU networks,” \emph{Neural Netw.}, vol. 94, pp. 103-114, Oct, 2017, doi: 10.1016/j.neunet.2017.07.002.
		
		\bibitem{b8} J. H. Cui, X. J. Liu, and T. Y. Chai, “Approximate scenario-based economic model predictive control with application to wind energy conversion system,” \emph{IEEE Trans. Ind. Inform.}, vol. 19, no. 4, pp. 103-114, Apr, 2023, doi: 10.1109/TII.2022.3189440.
		
		\bibitem{b9} J. P. Jordanou, E. A. Antonelo, and E. Camponogara, “Echo state networks for practical nonlinear model predictive control of unknown dynamic systems,” \emph{IEEE Trans. Neural Netw. Learn. Syst.}, vol. 33, no. 6, pp. 2615-2629, Jun, 2022, doi: 10.1109/TNNLS.2021.3136357.
		
		\bibitem{b10} P. Egli and H. Marco, “Towards RL-based hydraulic excavator automation,” in \emph{Proc. IEEE/RSJ Int. Conf. Intell. Robots Syst. (IROS)}, 2020, pp. 2692--2697.
		
		\bibitem{b11} S. B. Wang and J. Na, “Parameter estimation and adaptive control for servo mechanisms with friction compensation,” \emph{IEEE Trans. Ind. Inform.}, vol. 16, no. 11, pp. 6816-6825, Nov, 2020, doi: 10.1109/TII.2020.2971056.
		
		\bibitem{b12} Y. F. Lv, X. M. Ren, J. Y. Tian, and X. W. Zhao, “Inverse-model-based iterative learning control for unknown MIMO nonlinear system with neural network,” \emph{Neurocomputing}, vol. 519, pp. 187-193, Jan, 2023, doi: 10.1016/j.neucom.2022.11.040.
		
		\bibitem{b13} J. Park, B. Lee, S. Kang, P. Kim, and H. Kim, “Online learning control of hydraulic excavators based on echo-state networks,” \emph{IEEE Trans. Autom. Sci. Eng.}, vol. 14, no. 1, pp. 249-259, Jan, 2017, doi: 10.1109/TASE.2016.2582213.
		
		\bibitem{b14} S. Kim, B. Cho, S. Shin, J. Oh, J. Hwangbo, and H. Park, “Force control of a hydraulic actuator with a neural network inverse model,” \emph{IEEE Robot. Autom. Lett.}, vol. 6, no. 2, pp. 2814-2821, Apr, 2021, doi: 10.1109/LRA.2021.3062353.
		
		\bibitem{b15} R. R. Selmic and F. L. Lewis, “deadzone compensation in motion control systems using neural networks,” \emph{IEEE Trans. Autom. Control}, vol. 45, no. 4, pp. 602-613, Apr, 2000, doi: 10.1109/9.847098.
		
		\bibitem{b16} M. Lee, H. Choi, C. Kim, J. Moon, D. Kim, and D. Lee, “Precision motion control of robotized industrial hydraulic excavators via data-driven model inversion,” \emph{IEEE Robot. Autom. Lett.}, vol. 7, no. 2, pp. 1912-1919, Apr, 2022, doi: 10.1109/LRA.2022.3142389.
		
		\bibitem{b17} J. Weigand et al., “Hybrid data-driven modelling for inverse control of hydraulic excavators,” in \emph{Proc. IEEE/RSJ Int. Conf. Intell. Robots Syst. (IROS)}, 2021, pp. 2127-2134.
		
		\bibitem{b18} C. Xie and S. L. Chen, “A Physics-Guided Reversible Residual Neural Network Model: Applied to Build Forward and Inverse Models for Turntable Servo System,” \emph{IEEE Trans. Ind. Inform.}, vol. 19, no. 4, pp. 5882-5890, Apr, 2023, doi: 10.1109/TII.2022.3200668.
		
		\bibitem{b19} J. Y. Yao and W. X. Deng, “Active Disturbance Rejection Adaptive Control of Hydraulic Servo Systems,” \emph{IEEE Trans. Ind. Electron.}, vol. 64, no. 10, pp. 8023-8032, Oct, 2017, doi: 10.1109/TIE.2017.2694382.
		
		\bibitem{b20} Y. R. Ko and T. H. Kim, “Feedforward Plus Feedback Control of an Electro-Hydraulic Valve System Using a Proportional Control Valve,” \emph{Actuators}, vol. 9, no. 2, Jun, 2020, Art. no. 45, doi: 10.3390/act9020045.
		
		\bibitem{b21} L. Dinh, D. Krueger, and Y. Bengio, “NICE: Non-linear independent components estimation,” in \emph{Proc. Int. Conf. Learn. Representations (ICLR)}, May, 2015. [Online]. Available: http://arxiv.org/abs/1410.8516.
		
		\bibitem{b22} L. Dinh, J. Sohl-Dickstein, and S. Bengio, “Density estimation using real NVP,” in \emph{Proc. Int. Conf. Learn. Representations (ICLR)}, Apr, 2017. [Online]. Available: http://arxiv.org/abs/1605.08803.
		
		\bibitem{b23} M. MacKay, P. Vicol, J. Ba, and R. Grosse, “Reversible recurrent neural networks,” in \emph{Proc. Adv. Neural Info. Proc. Syst.}, Oct, 2018. [Online]. Available: http://arxiv.org/abs/1810.10999.
		
		\bibitem{b24} K. Khalil, B. Dey, A. Kumar and M. Bayoumi, “A reversible-logic based architecture for long short-term memory (LSTM) network,” in \emph{Proc. IEEE Int. Symp. Circuits Syst. (ISCAS)}, 2021, pp. 1-5.
		
		\bibitem{b25} D. Wang and J. Huang, “Neural network-based adaptive dynamic surface control for a class of uncertain nonlinear systems in strict-feedback form,” \emph{IEEE Trans. Neural Netw.}, vol. 16, no. 1, pp. 195–202, Jan, 2005, doi: 10.1109/TNN.2004.839354.
		
		\bibitem{b26} A. Grover, M. Dhar, and S. Ermon, “Flow-GAN: Combining maximum likelihood and adversarial learning in generative models,” in \emph{Proc. AAAI Conf. Artif. Intell.}, 2018, pp. 1-8.
		
		\bibitem{b27} W. X. Deng, J. Y. Yao, and D. W. Ma, “Robust adaptive precision motion control of hydraulic actuators with valve dead-zone compensation,” \emph{ISA Trans.}, vol. 70, pp. 269-278, Sep, 2017, doi: 10.1016/j.isatra.2017.07.022.
		
	\end{thebibliography}
\end{document}